\documentclass[acmsmall]{acmart}

\usepackage{booktabs}

\setcopyright{acmlicensed}
\acmJournal{PACMHCI}
\acmYear{2022} \acmVolume{6} \acmNumber{CSCW1} \acmArticle{52} \acmMonth{4} \acmPrice{15.00}\acmDOI{10.1145/3512899}

\begin{document}

\title[Assessing the Fairness of AI Systems]{Assessing the Fairness of AI Systems: AI Practitioners' Processes, Challenges, and Needs for Support}



\author{Michael Madaio}
\affiliation{%
  \institution{Microsoft Research}
  \city{New York City}
  \state{NY}
  \country{USA}}

\author{Lisa Egede}
\affiliation{%
  \institution{Carnegie Mellon University}
  \city{Pittsburgh}
  \state{PA}
  \country{USA}}

  \author{Hariharan Subramonyam}
\affiliation{%
  \institution{Stanford University}
  \city{Stanford}
  \state{CA}
  \country{USA}}

\author{Jennifer Wortman Vaughan}
\affiliation{%
  \institution{Microsoft Research}
  \city{New York City}
  \state{NY}
  \country{USA}}

  \author{Hanna Wallach}
\affiliation{%
  \institution{Microsoft Research}
  \city{New York City}
  \state{NY}
  \country{USA}}

\renewcommand{\shortauthors}{Madaio et al.}

\begin{abstract}
  Various tools and practices have been developed to support
  practitioners in identifying, assessing, and mitigating
  fairness-related harms caused by AI systems. However, prior research
  has highlighted gaps between the intended design of these tools and
  practices and their use within particular contexts, including gaps
  caused by the role that organizational factors play in shaping
  fairness work. In this paper, we investigate these gaps for one such
  practice: disaggregated evaluations of AI systems, intended to
  uncover performance disparities between demographic groups. By
  conducting semi-structured interviews and structured workshops with
  thirty-three AI practitioners from ten teams at three technology
  companies, we identify practitioners' processes, challenges, and
  needs for support when designing disaggregated evaluations. We find
  that practitioners face challenges when choosing performance
  metrics, identifying the most relevant direct stakeholders and
  demographic groups on which to focus, and collecting datasets with
  which to conduct disaggregated evaluations. More generally, we
  identify impacts on fairness work stemming from a lack of engagement
  with direct stakeholders or domain experts, business imperatives
  that prioritize customers over marginalized groups, and the drive to
  deploy AI systems at scale.\looseness=-1
\end{abstract}

\ccsdesc[500]{Human-centered computing~Empirical studies in collaborative and social computing}
\ccsdesc[500]{Human-centered computing~Collaborative and social computing design and evaluation methods}
\ccsdesc[500]{Computing methodologies~Artificial intelligence}


\keywords{AI; machine learning; fairness; software development practices}

\maketitle

\section{Introduction}

Artificial intelligence (AI) is now ubiquitous in both mundane and
high-stakes domains, including education, healthcare, and finance, yet
it is increasingly clear that AI systems can perform differently for
different groups of people, typically performing worse for those groups that are
already marginalized within
society~\cite[e.g.,][]{buolamwini2018gender, obermeyer2019dissecting,
  koenecke2020racial}. Too often, such performance disparities are
uncovered only after AI systems have been deployed and even after people
have already experienced fairness-related harms
\cite{Holstein:2019fr}.\looseness=-1

When the performance of an AI system is assessed in aggregate, poor performance for
particular groups of people can be obscured. Disaggregated evaluations
are intended to uncover such performance disparities by assessing
performance separately for different demographic groups.\footnote{We
  note that although disaggregation is most often done based on
  demographic groups such as race, gender, and socioeconomic status, it
  is possible to disaggregate performance by any set of
  groups.} Disaggregated evaluations have been the foundation of much
of the literature on assessing the fairness of AI systems
\cite[e.g.,][]{buolamwini2018gender,obermeyer2019dissecting,koenecke2020racial,ngan2020face},
including work such as the Gender Shades project
\cite{buolamwini2018gender}, which found differences in the
performance of three commercially available gender classifiers for
groups based on gender and skin tone, and related work from the
U.S. National Institute of Standards and Technology on differences in
the performance of face-based AI systems for groups based on sex, age,
race, and other factors \cite{ngan2020face}.  Disaggregated
evaluations have also been used to uncover race-based performance
disparities exhibited by healthcare systems
\cite{obermeyer2019dissecting} and commercially available
speech-to-text systems \cite{koenecke2020racial}.

To support practitioners in assessing the fairness of AI systems,
Barocas et al. articulated a set of choices, considerations, and
tradeoffs involved in designing disaggregated evaluations
\cite{barocas2021designing}, including why, when, by whom, and how
such evaluations should be designed and conducted, modeled after
similar approaches in other industries such as the U.S. Food and Drug
Administration's clinical trials. However, prior research suggests that practitioners may have difficulty
adapting fairness tools and practices for use in their contexts
\cite{Holstein:2019fr}. Moreover, tools and practices that do not align
with practitioners' workflows and organizational incentives may not be
used as intended or even used at all
\cite{madaio2020co,rakova2020responsible,lee2020landscape,richardson2021towards}. Although
researchers have studied how practitioners use fairness toolkits
\cite[e.g.,][]{lee2020landscape,richardson2021towards}, disaggregated
evaluations have not yet been studied within the
organizational contexts of practitioners working to
assess the fairness of the AI systems that they are developing. As a result, it is not clear how their
situated work practices influence disaggregated evaluations. To
address this, we ask three research questions:

\textbf{RQ1}: What are practitioners' existing processes and
challenges when designing disaggregated evaluations of
their AI systems?

\textbf{RQ2}: What organizational support do practitioners need when
designing disaggregated evaluations, and how do they communicate those
needs to their leadership?

\textbf{RQ3}: How are practitioners' processes, challenges, and needs
for support impacted by their organizational contexts?

To investigate these research questions, we conducted semi-structured
interviews and structured workshops intended to walk participants
through a process for designing disaggregated evaluations that we
adapted from \citet{barocas2021designing}. Thirty-three practitioners
took part in our study, from ten teams responsible for developing AI
products and services (e.g., text prediction systems, chatbot systems,
text summarization systems, fraud detection systems) at three
technology companies, in a variety of roles (e.g.,
program or product managers, data scientists, user experience
designers). We find that practitioners face challenges when designing
disaggregated evaluations, including challenges when choosing
performance metrics, identifying the most relevant direct
stakeholders\footnote{Direct stakeholders are the people that use or
  operate an AI system, as well as anyone that could be directly
  affected by someone else's use of the system, by choice or not. In
  some cases, customers are direct stakeholders, while in other cases
  they are not.\looseness=-1} and demographic groups on which to focus, and
collecting datasets with which to conduct disaggregated
evaluations. We also highlight how priorities for assessing the
fairness of AI systems may compound existing inequities, despite
practitioners' best intentions, and we identify tensions in
practitioners' desires for their organizations to provide guidance
about and resources for designing and conducting disaggregated
evaluations.\looseness=-1

This paper contributes to the growing literature on practitioners'
needs when operationalizing fairness in the context of AI system
development \cite[e.g.,][]{Holstein:2019fr} and to specific
conversations within CSCW and adjacent communities around anticipating
potential harms caused by sociotechnical systems
\cite[e.g.,][]{shilton2015anticipatory,brey2012anticipatory,boyarskaya2020overcoming},
including the role that organizational factors play in shaping
fairness work
\cite[e.g.,][]{madaio2020co,rakova2020responsible,metcalf2019owning,miceli2020between}. We
conclude by discussing some implications of our findings, including
the impacts of business imperatives (such as those that
prioritize customers over marginalized groups) on disaggregated
evaluations.
Finally, we discuss how the scale at which AI systems are deployed may impact
disaggregated evaluations due to a lack of situated knowledge of what
marginalization means in different geographic contexts.

\section{Related work}

\subsection{Assessing the fairness of AI systems}
Disaggregated evaluations are often designed and conducted by third
parties that are external to the teams responsible for developing the
AI systems to be evaluated.\footnote{When disaggregated evaluations
  are designed and conducted by third parties, they are often referred
  to as audits, but following Barocas et
  al. \cite{barocas2021designing}, we avoid the use of this term.}
Although there are good reasons for third parties to design and
conduct disaggregated evaluations, including increased credibility (as
there may be fewer reasons to believe that decisions were made so as
to cast the AI systems in question in a more favorable light), there are also
drawbacks to relying solely on this approach
\cite{barocas2021designing}. For example, third parties may not have access to
detailed knowledge about AI systems' inner workings, perhaps due to
trade secrecy laws \cite{katyal2019private}, and may therefore
overlook crucial components or features that might cause
fairness-related harms \cite{kroll2018fallacy}.
In addition, unless third parties are given pre-deployment access,
they can only conduct disaggregated evaluations of AI systems that
have already been deployed and potentially even caused
fairness-related harms \cite{raji2020closing}---or what Morley et
al. referred to as a gap between diagnostic and prescriptive
approaches \cite{morley2021ethics}.\looseness=-1

To address this, Raji et al. proposed the
SMACTR framework for internal auditing to close the ``AI
accountability gap'' \cite{raji2020closing}, with prompts and
artifacts targeted at different stages of the AI development
lifecycle. This framework is designed to promote internal auditing
practices that might align evaluations of AI systems with
organizations' principles and values. However, it is intended to be
used by auditors that are internal to the organizations responsible for
developing the AI systems in question, but external to the teams
tasked with system development. This raises the question of how
practitioners can use disaggregated evaluations to uncover performance
disparities before system deployment, particularly given that
incentives to ship AI products and services quickly may be at odds with
the slow and careful work that is needed to design and conduct such
evaluations \cite{madaio2020co,moss2021assembling}.\looseness=-1

\subsection{Anticipatory ethics in technology development}

Practices that are intended to uncover potential fairness-related
harms during the AI development lifecycle fit within a larger
tradition of research from the HCI, STS, and design communities on
anticipating potential harms, broadly construed, caused by
sociotechnical systems.\looseness=-1

This work has articulated how future-oriented discourses and practices
may act as ``anticipation work''
\cite[e.g.,][]{steinhardt2015anticipation, clarke2016anticipation}, or
the ``complex behaviors and practices that define, enact, and maintain
vision'' to ``define, orient, and accommodate to expectations of the
future'' \cite{steinhardt2015anticipation}. Steinhardt and Jackson
traced how individual and collective actors in oceanographic research
do the mundane work of wrangling rules, procedures, protocols, and
standards to anticipate potential downstream impacts of their
research, across multiple temporal and geographic scales
\cite{steinhardt2015anticipation}. Although their work was not focused
on AI systems or even technology, they discuss how anticipation work
makes normative claims about the kinds of futures that those actors intend
to bring about and, as such, is value-laden. Similarly, assessing the
fairness of AI systems during the development lifecycle requires
practitioners to perform anticipation work by considering potential
harms to different groups of people.\looseness=-1

Others have focused more explicitly on anticipatory ethics in
technology development
\cite[e.g.,][]{brey2012anticipatory,johnson2011software,shilton2015anticipatory,shilton2015s,floridi2020ethical}. Brey
outlined an approach called anticipatory technology ethics, pointing
out the uncertainty in identifying potential consequences of emerging
technologies during research and development, and arguing for the
importance of understanding use cases and deployment contexts for
future applications in order to identify potential harms
\cite{brey2012anticipatory}. Building off of this, Shilton conducted
ethnographic work to understand how technology developers put
anticipatory technology ethics into practice, highlighting the role of
``values levers'' in prying open (i.e., prompting) conversations about
particular ethical tensions in the development lifecycle.



In the AI community, the NeurIPS conference has required authors to
identify potential negative impacts of their research contributions
\cite{prunkl2021institutionalizing}.  This practice echoes recent
calls in the HCI community to remove the ``rose-tinted glasses'' of
researchers' optimistic visions for their research in order to
identify potential negative impacts \cite{H+18}.  Within AI,
Nanayakkara et al. offered a typology of uncertainty in anticipatory
ethics work \cite{nanayakkara2020anticipatory} and Boyarskaya et
al. conducted a survey of NeurIPS papers to understand AI researchers'
``failures of imagination'' in envisioning potential harms caused by
their work \cite{boyarskaya2020overcoming}. However, these approaches
to anticipatory ethics are often targeted at researchers (who may view
their work as more theoretical or conceptual than applied), rather
than at practitioners responsible for developing AI systems that will
be deployed. Moreover, such approaches often frame the challenge as
one for individuals, without grappling with the collaborative nature
of work practices, embedded within organizational contexts.

\subsection{Fairness work in organizational contexts}

AI practitioners, like other social actors, are embedded within
organizational contexts that shape the nature of their work practices,
and that can contribute to gaps between the
intended design of fairness tools and practices and their
use \cite[cf.][]{ackerman2000intellectual}. There is a long tradition
of research in the HCI and STS communities focusing on workplace
studies that situate technology work practices within their
organizational contexts \cite[e.g.,][]{schmidt2000critical,
  suchman2002located}. In recent years, considerable research in CSCW
has taken a situated approach to understanding how the work practices
of data scientists \cite{passi2018trust,muller2019human}, AI
developers \cite{piorkowski2021ai}, and software developers are
shaped by the organizational contexts within which they are
embedded
\cite{wolf2018changing,wolf2018participating,wolf2019conceptualizing,wolf2019evaluating,wolf2020making}. For
example, Passi and Sengers described collaborative, interdisciplinary
teams of practitioners that ``mak[e] data science systems work''
\cite{passi2020making}, identifying organizational factors that
empower business actors over other team members and business
imperatives that prioritize particular normative goals over other
goals.\looseness=-1

In the context of fairness, several recent papers have focused on how
organizational factors, including organizational cultures and
incentives, can impact practitioners' efforts to conduct fairness work
\cite[e.g.,][]{rakova2020responsible,madaio2020co,Holstein:2019fr,passi2019problem,hutchinson2021towards,miceli2020between,metcalf2019owning}. For
example, Madaio et al. highlighted how business imperatives to ship
products and services on rapid timelines were not aligned with the
timelines needed for fairness work, identifying perceived social costs
and risks to promotion for practitioners that choose to
raise fairness-related concerns \cite{madaio2020co}. Rakova et
al. found similar misaligned incentives for fairness work
\cite{rakova2020responsible}, identifying emerging trends for
practices such as proactive assessment and mitigation of
fairness-related harms (similar to the SMACTR framework proposed by
Raji et al. \cite{raji2020closing} and the fairness checklist proposed
by Madaio et al. \cite{madaio2020co}).

Organizational factors impact every aspect of fairness work, from
determining what fairness might mean for a particular AI system
\cite{passi2019problem} to dataset annotation, where dynamics between
annotators and other actors can impact annotators' labeling decisions
\cite{miceli2020between}, and beyond
\cite[e.g.,][]{madaio2020co,rakova2020responsible}. It is therefore
difficult to, as some have claimed, ``disentangle normative questions
of product design... from empirical questions of system
implementation'' \cite{bakalar2021fairness}, given the ways in which
practitioners' work practices are shaped by the organizational
contexts within which they are embedded
\cite[e.g.,][]{madaio2020co,miceli2020between,rakova2020responsible}.\looseness=-1

\section{Methods}

\subsection{Study design}
In order to investigate our research questions, we developed a
two-phase study protocol for working with teams of AI practitioners
over the course of three sessions (the second phase was divided into
two sessions due to time constraints).\footnote{We have included the protocol for the interviews and workshop sessions as supplementary material.} The study was approved by our
institution's IRB. Participation was voluntary and all participants were compensated for their participation. Phase one consisted of a 30--60-minute
semi-structured interview with a program or product manager (PM) on
each team that we recruited. We conducted these interviews to
understand teams' development practices for their AI systems and
their current fairness work, if any. Phase two consisted of two
90-minute structured workshop sessions with multiple members of each
PM's team. These workshop sessions were intended to help teams
design disaggregated evaluations of their AI systems. The time between
each team's two workshop sessions ranged from two to six weeks,
depending on participants' schedules. At the end of the first workshop
session, we encouraged participants to continue the design process
(i.e., to choose performance metrics, as well as to identify the most
relevant direct stakeholders and demographic groups on which to focus)
on their own time. At the start of the second workshop session, we
asked participants whether they had held any additional meetings to
discuss the topics covered in the first workshop
session. Although some teams described other ongoing fairness work,
none had discussed the topics covered in the first workshop session.\looseness=-1

\subsubsection{Participants}

We recruited participants working on AI products and services via a
combination of purposive and snowball sampling. We used both direct
emails and posts on message boards related to the fairness of AI
systems. We asked each participant to send our recruitment email to
other contacts and, in particular, to other members of their team in
roles different to theirs so that multiple team members could
participate in the workshop sessions together. We sought to recruit
participants working on a variety of AI systems, though in practice
many of our participants were members of teams developing language
technologies, as discussed in section~\ref{limitations}.\looseness=-1

Thirty-three practitioners took part in our study, from ten teams at
three technology companies of varying sizes, although only seven of
the ten teams completed both phases. Participants had a variety of
roles beyond PM, including technical roles (such as data scientist,
applied scientist, and software developer), and design roles (such as
user experience (UX) designer). Participating teams were
responsible for developing AI products and services related to text
suggestion, chatbots, text summarization, fraud detection, and
more. Table~\ref{tab:participants} contains more details about
participants and their teams. Specific details about their companies
and their products and services have been abstracted to preserve
anonymity, which was a condition of participation in the study.\looseness=-1

\begin{table}[]
\footnotesize
\begin{tabular}{lllll}
\toprule
  \textbf{Team} & \textbf{Product or service}         & \textbf{Phases} & \textbf{Participants}             & \textbf{Roles}        \\
  \midrule
1             & Resume skill identification & 1, 2a, 2b         & P1, P6, P9, P10, P11             & PM, UX, T, UX, T      \\
2             & Text prediction             & 1, 2a, 2b         & P3, P13, P14, P15                & UX/PM, T, PM, PM      \\
3             & Chatbot                     & 1, 2a, 2b         & P16, P18, P19                    & TM, T, T              \\
4             & Text summarization          & 1, 2a, 2b         & P7, P20, P21, P22, P29           & PM, TM, PM, PM, TM    \\
5             & Meeting summarization       & 1, 2a, 2b         & P8, P23, P24, P25, P26, P27, P28 & PM, PM, T, T, T, T, T \\
6             & Text clustering             & 1, 2a, 2b         & P30, P31                         & TM, T                 \\
7             & Fraud detection             & 1, 2a, 2b         & P32, P33                         & T, TM                 \\
8             & Text prediction             & 1               & P2, P17                          & PM, TM                \\
9             & Security threat detection   & 1               & P4                               & PM                    \\
10            & Speech to text              & 1               & P5                               & PM \\
\bottomrule
\end{tabular}
\caption{Participants and their teams. Roles included program or
  product manager (PM), user experience designer (UX), technical
  individual contributor roles (e.g., data scientist; T), and
  technical manager (TM).}
\label{tab:participants}
\end{table}

\subsubsection{Semi-structured interviews}

As described above, phase one of our study consisted of a
30--60-minute semi-structured interview with a PM on each team that we
recruited. Due to the COVID-19 pandemic, we conducted all interviews
remotely on a video conferencing platform. Our interview protocol was
informed by prior research on AI practitioners' fairness work
\cite[e.g.,][]{Holstein:2019fr,madaio2020co} and was designed to help
us investigate our first research question and plan the workshop
sessions in phase two. For example, we first asked each PM to describe
their team's development practices (e.g., ``Can you walk us through
the process for designing this product or service?'')  and to focus on
a particular component or feature of their AI system (e.g., ``Which
system component or feature would it make the most sense to focus on
for planning a disaggregated evaluation?''). Then, we asked questions
designed to help us understand their team's existing practices for
assessing the fairness of their AI system (e.g., ``Can you talk us
through any discussions or planning your team has done for identifying
or assessing fairness-related harms for your product or
service?''). At the beginning of the interview, the PM was given an
overview of the study and told what to expect in each of the sessions;
at the end of the interview, they were given the opportunity to
suggest other members of their team so that multiple team members
could participate in the workshop sessions together.\looseness=-1

\subsubsection{Structured workshops}

Phase two consisted of two 90-minute structured workshop sessions
(also conducted on a video conferencing platform) with multiple
members of each PM's team. Participants were provided with a slide
deck in which to document their responses and collectively brainstorm
ideas. Our goal for this phase was to understand how teams design
disaggregated evaluations of their AI systems, based on the process
proposed by Barocas et al. \cite{barocas2021designing}, described
below. At the end of each workshop session, we asked participants a
series of reflection questions, probing on what was challenging about
the process, what organizational support they might need, and how the
process might work in their organizational context. These reflection
questions were informed by our prior research and experiences with AI practitioners,
and were designed to help us investigate our second and third research
questions.

\subsection{Designing disaggregated evaluations}
\label{disaggregated_evaluations}

As described above, we used the workshop sessions to introduce
participants to a process for designing disaggregated evaluations,
based on the choices, considerations, and tradeoffs articulated by
Barocas et al. \cite{barocas2021designing}. This process consists of a
series of questions to be answered when designing a disaggregated
evaluation, including why, when, by whom, and how the evaluation
should be designed and conducted. We
determined answers to some of these questions in advance (e.g., the
evaluation was to be designed by our participants in collaboration
with one of the authors, who led the workshop sessions). Other
questions were answered during the interviews (e.g., the component or
feature on which to focus the evaluation), while the rest were
answered during the workshop sessions. Below, we explain the parts of this
process that are particularly important for our findings.\looseness=-1

At a high level, the protocol for the workshop sessions consisted of
prompts for each team to decide 1) what good performance meant for
their AI system and how this might be operationalized via performance
metrics, 2) which use cases and deployment contexts to consider, 3)
which direct stakeholders and demographic groups would be most
relevant to focus on, 4) what data would therefore be needed to
conduct the disaggregated evaluation, and 5) how best to determine
whether their system was behaving fairly using the chosen
performance metrics, the identified direct stakeholders and demographic groups, and the proposed dataset. As most of our findings relate to the
first, third, and fourth decisions, we provide more details about each
of these below.\looseness=-1

\subsubsection{Choosing performance metrics.}

We first asked each team what good performance meant for their AI
system, broadly construed. Then, we asked them to anchor this
definition in specific performance metrics that could be used to
conduct their disaggregated evaluation. To do this, we asked them
whether there were performance metrics that they already used to
assess the performance of their AI system, and then probed on what
those metrics might miss, including any aspects of performance
that may be more likely to exhibit disparities indicative of
fairness-related harms. Although some teams identified standard
performance metrics (e.g., word error rate for speech-to-text
systems), others did not. We asked each team to revisit their list of
performance metrics throughout both workshop sessions (adding new
metrics as appropriate) and, ultimately, to choose one or more metrics
on which to focus their disaggregated evaluation.\looseness=-1

\subsubsection{Identifying direct stakeholders and demographic groups}
Having asked each team to decide which use cases and deployment
contexts to consider, we then asked them to identify the direct
stakeholders---that is, anyone that would use or operate their system,
as well as anyone that could be directly affected by someone else's use
of the system, by choice or not---that they thought were most likely
to experience poor performance. For each of the identified direct
stakeholders, we further asked which demographic groups might be most
at risk of experiencing poor performance. We did this by providing
examples of different demographic factors (e.g., race, age, gender)
and asking participants to identify the demographic factors that they
thought would be most relevant to consider. Then, for each factor, we
asked them to identify the groups that they wanted to focus on. For
example, having decided to consider age, a team might choose to focus
on three age groups: under eighteen, eighteen to sixty-five, and over
sixty-five.

\subsubsection{Collecting datasets}
Conducting a disaggregated evaluation requires access to an
appropriate dataset with sufficient data points from each demographic
group of interest, labeled so as to indicate group membership. In
addition, the dataset must also capture realistic within-group
variation and the data points must be labeled with relevant
information about any other factors (e.g., lighting conditions) that
might affect performance. We asked each team whether they
already had access to an appropriate dataset with which to conduct
their disaggregated evaluation or whether they would need to collect a
new dataset. If a team said that they would need to collect a new
dataset, we asked them how they might go about doing this.

\subsection{Data analysis}

To understand the most salient themes in the transcripts from our
interviews and workshop sessions, we adopted an inductive thematic
analysis approach \cite{braun2006using}. Three of the authors
conducted an open coding of the transcripts using the qualitative
coding software Atlas.ti. They first coded the same transcript and
discussed codes. After that, they then divided up the remaining
transcripts among themselves and iteratively grouped the codes into a
set of themes using virtual whiteboard software. As part of this
iterative sense-making process, all five authors discussed the
emerging themes, and three of the authors re-organized the codes
(merging redundant codes as needed) and structured the themes (i.e.,
into higher and lower levels) multiple times with input from the other
authors. Throughout the coding process, two of the authors wrote
reflective memos~\cite[e.g.,][]{birks2008memoing,lempert2010asking} to
ask analytic questions of emerging thematic categories and to make
connections across themes and across the phases of the study. These
memos were informed by the authors' experiences conducting the
interviews and workshop sessions, as well as by artifacts from the
workshop sessions, including the notes that the authors took while
conducting the sessions and the notes that participants took on
the slide decks. In section~\ref{sec:findings}, we discuss the
findings identified from these codes, themes, and reflective memos.\looseness=-1

\subsection{Positionality}
In the interest of reflexivity \cite{braun2019reflecting,
  liang2021embracing}, we acknowledge that our perspectives and
approaches to research are shaped by our own experiences and
positionality. Specifically, we are researchers living and working in
the U.S., primarily working in industry\footnote{Two of the authors
  are in academia, but they were interns in industry when we conducted
  the study.} with years of experience working closely with AI
practitioners on projects related to the fairness of AI systems. In
addition, we come from a mix of disciplinary backgrounds, including AI
and HCI, which we have drawn on to conduct prior research into
sociotechnical approaches to identifying, assessing, and mitigating
fairness-related harms caused by AI systems.

\section{Findings}
\label{sec:findings}

We find that practitioners face challenges when designing
disaggregated evaluations of AI systems, including challenges when
choosing performance metrics, identifying the most relevant direct
stakeholders and demographic groups on which to focus (due to a lack
of engagement with direct stakeholders or domain experts), and
collecting datasets with which to conduct disaggregated
evaluations. We discuss how the heuristics that teams use to determine
priorities for assessing the fairness of AI systems are shaped by
business imperatives in ways that may compound existing inequities,
despite practitioners' best intentions. Finally, we cover
practitioners' needs for organizational support, including guidance on
identifying direct stakeholders and demographic groups and strategies
for collecting datasets, as well as tensions in organizational
processes for advocating for resources for designing and conducting
disaggregated evaluations.\looseness=-1

\subsection{Challenges when designing disaggregated evaluations}
\label{challenges}

Below we describe some of the challenges surfaced by participants
during the interviews and workshop sessions. As described above, the
most salient themes relate to choosing performance metrics,
identifying direct stakeholders and demographic groups, and collecting
datasets with which to conduct disaggregated evaluations. In the
absence of engagement with direct stakeholders or domain experts and
in the absence of opportunities for data collection, participants
described drawing on their own experiences and using their own
data---approaches that may impact teams' abilities to effectively
assess fairness-related harms experienced by people that do not
resemble AI practitioners.\looseness=-1

\subsubsection{Challenges when choosing performance metrics}
\label{metrics}

For some teams, choosing performance metrics was straightforward
because they decided to use the same performance metrics that they
already used to assess the aggregate performance of their AI systems. Some teams
even noted that there were standard performance metrics for AI systems
like theirs (e.g., word error rate for speech-to-text systems), making
their decisions relatively easy
\cite[cf.][]{buolamwini2018gender,koenecke2020racial}. However, most
teams did not have standard performance metrics. Choosing performance
metrics was therefore a non-trivial task, requiring lengthy
discussions during the workshop sessions about what good performance
meant for their AI systems, how aggregate performance was typically assessed, and whether or how this should change when designing a
disaggregated evaluation.

In these cases, participants described how their typical approaches to
assessing aggregate performance included metrics that they felt were
inappropriate to use when assessing the fairness of their AI systems. For
example, a UX PM on a team developing a text prediction system noted
that their usual performance metrics focused on usage (i.e., the rate
of acceptance of text predictions), but that they thought these
metrics did not capture aspects of performance that may be more likely to exhibit disparities indicative of
fairness-related harms. They told us,
\textit{``Today, our features are often judged by usage, but usage is
  not a success metric. Similarly, precision or recall, these model
  metrics are important, but they don't tell us whether we're actually
  achieving the outcomes that we want to achieve''} (P3, T2). For this
team, differences in usage between demographic groups would not
necessarily reveal anything about fairness-related harms, although
such differences might be suggestive of other differences that, in turn, might
reveal something about fairness-related harms.

To address this, some participants described how their teams were
developing new performance metrics specifically for assessing
the fairness of their AI systems, saying, for example, \textit{``We need to start quantifying
  and optimizing towards not just success, but output quality and how
  catastrophic failures are. So we invented new metrics around
  quality''} (P3, T2). This team therefore wanted to use this new metric for their disaggregated evaluation. However, other teams that created new performance metrics had difficulty agreeing
on whether they should use these new metrics for their disaggregated
evaluations or whether they should use the metrics that they typically used to assess performance (T1, T2,
T3, T5).

More generally, we find that decisions about performance metrics are
shaped by business imperatives that prioritize some stakeholders
(e.g., customers) over others (e.g., marginalized
groups). Participants described how tensions that arose during
discussions about choosing performance metrics---even for teams that
had standard performance metrics (e.g., word error rate)---were often
indicative of deeper disagreements about the goals of their
AI systems. These tensions were exacerbated by a lack of engagement with
direct stakeholders or domain experts, which we discuss in
section~\ref{stakeholder_identification}.\looseness=-1

For example, participants on a team developing a fraud detection
system (T7) described \textit{``a lot of tradeoffs in the project''} (P32, T7)
between three primary types of stakeholders: the people whose
transactions might be mistakenly labeled as fraudulent, the companies
running the money-transfer platforms on which the fraud detection
system was deployed, and local government fraud auditors that audit
the money-transfer platforms' transactions. They told us how
\textit{``it came close to becoming a literal fight because fraud
  detection wanted a model as strict as possible and they wanted for
  us to focus on that metric, while the business part of the model
  wanted it to be more flexible [...] and didn't want to classify
  almost every client [as fraudulent]''} (P32, T7). These tensions
between stakeholders with very different goals shaped the
team's discussion about choosing performance metrics. As they told
us:\looseness=-1

\begin{quote}
\textit{In a lot of cases we are able to [...] just take both
  requirements and get to some agreement. This was not exactly such a
  case, because I think it was a deeper conflict in the project itself
  [...] at the end we got [both stakeholders] to sit down together and
  we basically told them to make this work. So after that it was a bit
  easier, the tension was still in there, but at least we could work}.
(P32, T7)
\end{quote}

Although P32 described being able to get two types of stakeholders to
\textit{``sit down together''} to \textit{``get to some agreement''}
about assessing the aggregate performance of their AI system, that conversation was
not focused on fairness. Therefore, when discussing how to assess the fairness of their AI system, they found
themselves unable to resolve the decision of which performance metrics
to use without consulting their stakeholders. Moreover, the conversation that P32 described did not involve any \textit{direct}
stakeholders---that is, the people whose transactions might be
mistakenly labeled as fraudulent---a challenge commonly raised by
participants, as discussed in the next section.\looseness=-1

The technical manager on the team articulated the decision to
prioritize customers (where customers means the companies
running the money-transfer platforms on which the fraud detection
system was deployed, not the people whose
transactions might be mistakenly labeled as fraudulent) succinctly,
saying that it wasn't the team's role
to define the metric, but \textit{``in our case the customer defines
  what is the best metric''} (P33, T7)---a situation that was exemplified when the
other team member told us how they \textit{``gave in to the commercial
  point of view''} (P32, T7). Decisions about performance metrics, as
with other decisions made when assessing the fairness of AI systems,
are therefore not value-neutral. Rather, they are shaped by the
organizational contexts within which practitioners are embedded, including
business imperatives.\looseness=-1

\subsubsection{Challenges when identifying direct stakeholders and demographic groups}
\label{stakeholder_identification}
Participants wanted to identify the direct stakeholders and
demographic groups that might be most at risk of experiencing poor
performance based on an understanding of what marginalization means
for different demographic groups in the geographic contexts where
their AI systems were deployed, and they wanted to do so by engaging
with direct stakeholders and domain experts. For example, participants described how they wanted to engage with domain experts that had experience studying or working with different demographic
groups, saying, \textit{``I think that that's where we do need to
  bridge the people who are experts on this and know the processes we
  should be going through before we [...] decide on implementations
  with our opinions of what is important and who is harmed''} (P15,
T2).  This participant went on to say, \textit{``For gender non-binary [...] We
  need to ensure we have the right people in the room who are experts
  on these harms and/or can provide authentic perspectives from lived
  experiences [...]  I think the same could be said about speakers of
  underrepresented dialects as well''} (P15, T2). Other participants
were explicit about the importance of engaging with experiential
experts (i.e., people with relevant lived experiences) in addition to
other domain experts, saying, \textit{``I don't know what community
  members that speak that sociolect would want. But [designers'
    decisions] should agree with a native speaker of the chosen
  language''} (P7, T4).

Despite their desires to engage with direct stakeholders and domain
experts, participants described how their typical development
practices usually only involved customers and users, and not other
direct stakeholders. Moreover, many teams only engaged with users when
requesting feedback on their AI systems (e.g., via user testing) instead
of engaging with users to inform their understanding of what
marginalization means for different demographic groups or to inform
other decisions made when designing disaggregated evaluations.

In the absence of processes for engaging with direct stakeholders or domain experts,
participants suggested drawing on the personal experiences and
identities represented on their teams to identify the most relevant
direct stakeholders and demographic groups on which to focus. However,
this approach is problematic given the homogeneous demographics of
many AI teams \cite{west2019discriminating}. A PM on a team developing
a text prediction system described the situation as follows:
\begin{quote}
    \textit{The problems we identify are very much guided by the
      perspectives of the people on the team who are working on and
      thinking about these issues. In that sense I think there could
      be a disconnect between the scope of the harms the customers may
      be experiencing; and the scope of the harms our team is trying
      to identify, measure, and mitigate.} (P15, T2)
\end{quote}

For many teams, challenges around engaging with direct stakeholders
or domain experts were rooted in organizational incentives for rapid
deployment to new geographic contexts. As one product
manager put it:
\begin{quote}
    \textit{We don't have the luxury of saying, `Oh, we are supporting
      this particular locale and this particular language in this
      particular circumstance.' No, no, no, we're doing it all! We're
      doing it all at once, and we are being asked to ship more
      faster. I mean, that is the pressure [...] and there will be
      tension for anything that slows that trajectory, because the gas
      pedal is to the metal.} (P7, T4)
\end{quote}

As this participant described, incentives for deploying AI systems
to new geographic contexts can lead to work practices that obscure or
exacerbate, rather than mitigate, fairness-related harms. Below, we
unpack these incentives further, including how they affect data
collection, priorities for assessing the fairness of AI systems, and
access to resources for designing and conducting disaggregated
evaluations.\looseness=-1

\subsubsection{Challenges when collecting datasets}
\label{data_collection}

Participants described challenges when collecting datasets with which
to conduct disaggregated evaluations, including tensions between their
organizations' privacy requirements and the need for demographic data
with which to disaggregate performance. One participant highlighted this gap by saying, \textit{``So I guess
  I'm just having trouble getting over the hurdle that I don't think
  we have a real approved data collection method [for data that lets
    us evaluate fairness] at all''} (P9, T1). They recognized the
restrictions that privacy requirements place on their fairness work as
\textit{``the competing interests between not collecting [personally
    identifiable] data in order to protect privacy, and having the
  right set of data that we've collected that is representative''}
(P9, T1). For this team, as well as others, the importance of
protecting personally identifiable information restricted their access
to demographic data---a tension
identified across multiple domains in prior research
\cite{bogen2020awareness, andrus2021we}. These \textit{``competing
  interests''} (P9, T1) place teams in a difficult bind, as their
organizations' privacy requirements stand at at odds with increasing
demands to assess the fairness of their AI systems.

Participants also reported a lack of expertise in collecting
demographic data:
\begin{quote}
    \textit{We don't have any existing precedent or framework for
      doing this. Right now it's completely exploratory like okay, we
      need demographic information in order to build fairer test sets
      and evaluate the potential for quality of service differences
      [...] should we pay to have a focus group or, you know, collect
      the data set in some way like that? And that's the level that
      it's at right now as far as I know, so, nowhere near scaling to
      other teams and other services to get us that information.}
    (P13, T2)
\end{quote}

This participant, an applied scientist on a team developing a text
prediction system, was grappling with a lack of existing methods,
frameworks, or practices that they could draw on when collecting
demographic data. Many teams had little experience collecting
demographic data in general, while others lacked specific expertise in
collecting demographic data for their use cases or deployment
contexts. For some teams (e.g., T1), this was due to existing
agreements with customers that meant that they would need to make
substantial modifications to their engineering infrastructure or
data-sharing policies. Participants shared experiences trying to
develop a \textit{``side channel data collection method''} (P9, T1),
discussing exploratory approaches to data collection that might allow
them to adhere to their organizations' privacy requirements, while
still yielding demographic data with which to disaggregate
performance. Participants told us that \textit{``we barely have access
  to datasets to begin with, so we take anything that we can get
  basically''} (P29, T4).

In the absence of processes for collecting appropriate datasets,
participants shared how they had used or might use data from their
teams or organizations. Multiple teams discussed this approach (T1,
T2, T5), with one participant telling us, \textit{``One thing we're
  exploring now, at the very early stages of exploration, is seeing at
  least if we can start gathering some of this demographic information
  that internal [company] employees are sharing with us''} (P3,
T2). Others on their team wondered about this approach, asking,
\textit{``There's a question of should we try to do this internally
  and ask people if they'd be willing to voluntarily share and
  associate demographic information with some of their communication
  data?''} (P13, T2). On other teams, participants told us,
\textit{``Until [privacy issues] get reconciled, the only path forward
  that we have at the moment is to [...] send out a request to some
  fraction of [our company's] employees, and then ask them if they'd
  agree to provide demographic data''} (P24, T5).\looseness=-1

However, much like drawing on the personal experiences and identities
represented on teams to identify the most relevant direct stakeholders
and demographic groups (see section~\ref{stakeholder_identification}),
this approach is problematic. It is unlikely that the demographics of
AI teams match the demographics of the direct stakeholders encountered
in their uses cases or deployment contexts. One PM told us, \textit{``Our
  hardest problem with regards to fairness is access to data to
  actually assess the fairness of the algorithm in a context that
  would actually look like [our deployment context]''} (P9, T1). As
with the challenges described in
section~\ref{stakeholder_identification}, this challenge was made
worse by organizational incentives for \textit{``pedal [...] to the metal''}
deployment to new geographic contexts (P7, T4).

Participants from one team (T5) pointed out that data from their team
would likely be English language data, which would not reflect the
linguistic diversity of their deployment contexts. Other participants
concurred, pointing out that such data would only include
\textit{``people who type like us, talk like us as the people who are
  building the systems''} (P13, T2), and would not be representative
of the people most likely to experience poor performance, especially
when business imperatives motivated deploying their AI systems to new
geographic contexts. Although deployment expansion could potentially
motivate discussions about disaggregated evaluations, including data
collection, many teams instead felt pressures to deploy before
assessing the fairness of their AI systems.

\subsection{Priorities that compound existing inequities}
\label{priorities}

During the workshop sessions, each team identified many more direct
stakeholders than could be the focus of a single disaggregated
evaluation (particularly given the variety of use cases and deployment
contexts that they wanted to consider). As a result, we asked
participants to discuss how they would prioritize these direct
stakeholders, as well as how they would prioritize different
demographic groups. Participants prioritized direct stakeholders and
demographic groups based on the perceived severities of fairness-related
harms, the perceived ease of data collection or of mitigating
performance disparities, the perceived public relations (PR) or brand
impacts of fairness-related harms on their organizations, and the
needs of customers or markets---all heuristics that may compound
existing inequities. Although we report our findings here as
high-level themes, there was little agreement about priorities during
the workshop sessions. As such, this was the part of our protocol that
elicited the most back-and-forth discussion.

\subsubsection{Perceived severities of fairness-related harms}
\label{priorities_severities}

Several participants wanted to prioritize direct stakeholders and
demographic groups based on the severities of fairness-related harms,
saying, \textit{``So, severity level?''} (P1, T1) and \textit{``I
  guess we could think of some measure of how bad it is for something
  to go wrong''} (P9, T1). Some teams
discussed how they were not able to directly measure the severities of
fairness-related harms caused by performance disparities, and would
therefore need to use the \textit{``perceived severity of the harm''} (P15, T2) as
a proxy. Indeed, as described in
section~\ref{stakeholder_identification}, many teams had no processes
for engaging with direct stakeholders to understand their experiences
with fairness-related harms.\looseness=-1

Participants grappled with how to operationalize the perceived
severities of fairness-related harms, with one applied scientist saying
\textit{``I think the scale of the impact is also very difficult to
  know without measuring it first''} (P19, T3), and another thinking
out loud that they would need to \textit{``kind of multiply that
  [harm] in a way by the percent or perceived breadth of users, the
  estimated amount, percent of users who fall within that group''}
(P13, T2). Using such quantified approaches, teams discussed
prioritizing different direct stakeholders and demographic groups in
terms of tradeoffs between benefits and harms, based on the perceived severities
of the harms or the number of people likely to experience them. Some
participants articulated the importance of understanding the impact of
multiple microaggressions on marginalized groups, saying that they
needed to identify \textit{``the accumulated harm to that demographic
  or that group of people''} (P13, T2), while other participants
thought about this in comparative terms, saying:\looseness=-1
\begin{quote}
    \textit{Either there's a potential to really help a group, like, a
      lot more than another, for example, or to cause more harm to a
      group, in which case you would want to prioritize minimizing
      that harm or if there's an opportunity to do a lot of good for
      one group and a little good for another group, that maybe you
      should prioritize the one that has the biggest impact.} (P19,
    T3)\looseness=-1
\end{quote}

The idea that harms can be quantified and compared with benefits
belies the reality of AI systems, which have complex, messy impacts on
a wide range of direct and indirect stakeholders that may experience
harms and benefits in a multiplicity of hard-to-measure ways
(cf. \citet{jacobs2021measurement} on the difficulties of measuring
constructs, including fairness, in the context of AI
systems). Moreover, quantified approaches to prioritizing direct
stakeholders and demographic groups may end up being majoritarian,
where larger groups are prioritized over smaller, but perhaps more
marginalized, groups. Grappling with this tension, one
participant wondered, \textit{``How big is this group?  Like, what's
  the size of the user base in this group?  So if it's like a very
  severe problem for a small group of users, versus a less severe
  problem for a large group of users. We have to think about, how do
  we measure that and how do we balance those two?''} (P23,
T5). However, given teams' challenges around engaging with direct
stakeholders or domain experts (see section~\ref{challenges}), prioritization decisions
based on teams' perceived severities of fairness-related harms may lead
to disaggregated evaluations that do not focus on the most relevant
direct stakeholders and demographic groups.

\subsubsection{Perceived ease of data collection or of mitigating performance disparities}
\label{priorities_data_collection}
Participants on multiple teams
(e.g., T1, T5) described how they would prioritize direct stakeholders
and demographic groups for which they already had data or with whom
they were already working in some capacity, saying, \textit{``One that
  comes to mind is that we're already working with is [the
    accessibility group], right. So people who are going to dictate
  their resume [...] might be disadvantaged. So I feel like that's the
  highest priority''} (P1, T1).\looseness=-1

In addition to prioritizing direct stakeholders and demographic groups
based on the perceived ease of data collection, teams also discussed making
prioritization decisions based on the perceived ease of mitigating
performance disparities:\footnote{We note that the workshop sessions were focused
  entirely on designing disaggregated evaluations and therefore did
  not cover mitigation of performance disparities; see
  section~\ref{disaggregated_evaluations} for more details about the
  protocol for the workshop sessions.}
\begin{quote}
\textit{I think another criteria that we sometimes use for
  prioritizing or addressing these things, is ease of mitigation
  implementation. So I know for British English [...] that was
  something that we could pretty quickly and easily address.} (P15,
T2)
\end{quote}
This was a popular approach, discussed by three of the seven teams
(T1, T2, T5). However, implicit in P15's point (which they grappled
with later on in the workshop sessions) is the fact that although poor
performance for British English speakers may be relatively easy to
mitigate, it is unlikely that British English speakers experience
especially bad performance, nor are they a group that is typically
marginalized by AI systems or more generally within society.

\subsubsection{Perceived PR or brand impacts}
\label{priorities_PR}
Many teams
asked what it would mean for their organization if particular direct
stakeholders or demographic groups were found to experience poor
performance. For example, continuing the conversation about poor
performance for British English speakers, one participant asked,
\textit{``Let's say that we had a low quality of service for British
  English speakers. Would that have a high, medium, or low impact,
  like PR impact on the company?''} (P13, T2). They went on to say:\looseness=-1
\begin{quote}
    \textit{The PR impact is in some ways loosely tied to the harms to
      individuals or groups, plus some aspects of what goes viral. I
      would venture to put African-American language speakers in a
      higher bucket for both of these. So maybe high on severity of
      harm to groups or individuals, and medium or high for PR
      impact.} (P13, T2)
\end{quote}

This participant discussed prioritizing direct stakeholders and
demographic groups based on the perceived severities of fairness-related
harms, as described above, as well as the potential for performance
disparities to go viral, leading to negative PR impacts, aligned
with other discussions around ``\textit{the potential for harm to the
  company}'' (P13, T2). Similarly, a software developer on a team
developing a chatbot system told us, \textit{``If a user interacts
  with the bot and finds something that is really offensive maybe they
  will post it on social media in the hopes that it might go viral. So
  we're trying to detect these things hopefully before they go
  viral''} (P18, T3). However, this approach prioritizes direct
stakeholders and demographic groups that use social media and that are
able to widely publicize their
concerns.\footnote{https://www.technologyreview.com/2021/07/13/1028401/tiktok-censorship-mistakes-glitches-apologies-endless-cycle/}

Some participants described priorities based on perceived brand
impacts, with one participant saying \textit{``If you think of this as
  a headline in a newspaper, that could be a way that we kind of think
  about some of these''} (P15, T2). Indeed, imagining newspaper headlines is a common way to think about fairness-related harms or values in technology
development more generally \cite[cf.][]{wong2021timelines}. Other
participants drew on high-profile failures of AI systems in the news
to inform their priorities. Participants cited the \textit{``classic
  proxy bias example from the Amazon resume system''} (P1, T1) and
other examples of fairness-related harms caused by competitors' AI
systems (e.g., T2, T5, T8).\looseness=-1

Participants on other teams shared similar views, saying, \textit{``As
  a brand, [our company] does not want to associate with those
  [performance disparities] [...] the biggest risk that the feature
  has is like when you see [performance disparities], the company
  brand is associated with it''} (P2, T8). However, this approach of
making prioritization decisions based on brand impacts is a tactic that may ignore concerns about direct stakeholders'
experiences in order to uphold an organization's image
\cite{meingast2007embedded}. One participant outlined how they
navigate this, saying:
\begin{quote}
\textit{We just have to strategically pick like, `hey, we think this
  is socially important right now; especially when it comes to a like
  gap in data, right? So, if there is one or two groups of people that
  we think are strategically important from a social perspective, and
  are likely to be underrepresented, or have different performance
  across multiple types of models.} (P7, T4)\looseness=-1
\end{quote}

These approaches are inherently backward looking, or \textit{``reflexive and
reactive''} (P13, T2), as they focus on performance disparities that
are uncovered after AI systems have been deployed. In contrast, reflexive
design (or ``reflection in action'' \cite{schon1983reflective})
approaches encourage practitioners to reflect on values and impacts
during the development lifecycle \cite{sengers2005reflective}.

\subsubsection{Needs of customers or markets}
\label{priorities_business}

Finally, participants described prioritizing direct stakeholders and
demographic groups based on the needs of customers and markets,
thereby shaping disaggregated evaluations in ways that may compound
existing inequities by reifying the social structures that lead to
performance disparities in the first place.  As one participant put it,
\textit{``If the organization is talking only about business
  performance, then the project is going to prioritize the groups that
  are the most important for the business''} (P32, T7). In section
\ref{metrics}, we similarly discussed the ways that decisions about
performance metrics are shaped by business imperatives that prioritize
customers over marginalized groups. Participants on other teams shared
similar sentiments, saying, \textit{``I'm also cognizant that business
  stakeholders are often very focused on business outcomes in terms of
  dollar value rather than fairness or something else which is a
  `good-to-have'\hspace{0pt}''} (P30, T6).

Multiple participants told us that their organizations were focused on their customers and that this affected their engagement with direct
stakeholders, in turn affecting prioritization. One PM said:\looseness=-1
\begin{quote}
     \textit{We keep saying that we are `customer obsessed,' right?
       But we shouldn't just be saying, `Oh, ninety-nine percent of our
       customers are white men who work in corporate America and we're
       going to talk to them and see what feature they want.'} (P3,
     T2)
\end{quote}

This participant articulated the concern that their primary customers
were members of demographic groups that were already over-represented
and privileged in the geographic contexts in which their AI system was
deployed, thereby unintentionally reifying existing social
structures. Much like the majoritarian approaches described in
section~\ref{priorities_severities}, prioritizing the needs of
high-value customers might mean prioritizing the most powerful direct
stakeholders and demographic groups that are less likely to be
marginalized.\looseness=-1

Participants on almost every team shared that their organizations used
strategic market tiers to prioritize deployment to new
geographic contexts. Indeed, many companies whose AI products and services are deployed around the world use tier-based deployment
approaches \cite[cf.][]{westland2002cost}. As one PM shared:\looseness=-1
\begin{quote}
    \textit{We also need to think about internationalization. So
      scaling [this system] across markets is another thing that we
      look at. Often we start our language-based initiatives for the
      North American market. So for Canada and the United States. For
      the English language. And then we have tiers of language like
      Tier 1 languages, Tier 2 languages, Tier 3. Mostly that is from
      a business opportunity point of view. And then we scale.} (P3,
    T2)
\end{quote}

In theory, tier-based deployment approaches might help teams focus
their fairness work on a single geographic context\footnote{P3 went on
  to explain that although \textit{``market doesn't mean language,''}
  strategic market tiers were often developed with geographic contexts
  in mind---an approach that may not map cleanly onto
  languages when applied to language technologies.\looseness=-1} at a time,
enabling them to learn from one context to the next. However, in
practice, participants reported that their AI systems' deployments did
not appear to depend on disaggregated evaluations, but were instead
aligned with market strategies and opportunities. One technical
manager described the situation, saying:
\begin{quote}
    \textit{Sometimes you have a very clear market signal that you need to go into a certain domain, or a certain language, but sometimes the fact that you don't have any signal doesn't mean that you don't need to invest in adding a new language [...]
    I think it's much more like market opportunity in this space [...]
    So, it's opportunistic, some of it is strategic in nature.} (P29, T4)\looseness=-1
\end{quote}

As this participant described, for many teams, decisions about where
to deploy their AI systems were sometimes strategic, based on \textit{``market
opportunity''} (P29, T4), and sometimes opportunistic, based on customer feedback
(thereby raising questions about the impacts of prioritizing
customers' needs, as discussed above) or other factors. Such
deployment approaches may compound existing inequities across
geographic contexts:\looseness=-1
\begin{quote}
    \textit{[Our system] is helping people be more productive. And if
      English-speaking languages, or tier 1 markets, are already ahead
      of the other markets in productivity, then us starting with
      these folks-- and let's say a tier 3 language getting a feature
      that makes people more productive two years later means that
      opportunity gap will keep widening and widening, right? So there
      is a ethical aspect here that is not necessarily programmed into
      our business growth objectives.} (P3, T2)
\end{quote}

Assuming that this team's AI system will indeed help people to be more
productive and that being more productive is a good thing, then, as P3 argued, the gap in opportunity between
lower-tier and higher-tier markets will widen. Prioritizing direct
stakeholders and demographic groups based on the needs of customers or
markets may similarly compound existing inequities.\looseness=-1

\subsection{Needs for organizational support}

Participants wanted their organizations to provide guidance about and
resources for designing and conducting disaggregated evaluations. In
particular, participants voiced the need for guidance on identifying
the most relevant direct stakeholders and demographic groups on which
to focus, as well as strategies for collecting datasets. Below, we
identify tensions in participants' desired organizational support, as
well as tensions in organizational processes for advocating for
resources for designing and conducting disaggregated evaluations.

\subsubsection{Guidance on identifying direct stakeholders and strategies for collecting datasets}

Participants wanted guidance from their organizations on identifying
the most relevant direct stakeholders and demographic groups on which
to focus their disaggregated evaluations:
\begin{quote}
    \textit{I would love to see coming top-down is [...] what are those different [demographic] factors and groups within each factor? [...] There is almost like an infinite list of factors... We need to be told like a company-wide list. Throw all the linguists and all the researchers at it and be, like, these are the list of [demographic] factors
    who we definitely want to guarantee fair and inclusive high quality of service for.} (P13, T2)
\end{quote}

Participants recognized that the challenges they were facing were not
specific to their AI systems, and were likely shared by other teams in
their organizations. As a result, they noted that organization-wide
guidance would help many teams all grappling with the same
challenges. They told us that \textit{``none of these are
  [product]-specific problems''} (P3, T2), and that they felt they
were \textit{``doing a lot of -- not reinventing, but
  \textit{inventing} of the wheel''} (P3, T2) and wanted their
organizations to better support them in their fairness work. In many
cases, participants felt that their teams didn't have expertise in the
languages that their AI systems supported across the geographic contexts
in which they were deployed, saying, \textit{``We don't have
  any resources around what kind of harm language can generate [...]
  Most of these languages I don't even have expertise on, right?''}
(P3, T2).

Participants also wanted guidance on the types of fairness-related
harms that might be caused by their AI systems. They may have requested
this guidance due to their teams' lack of processes for engaging with
direct stakeholders or domain experts, as discussed in
section~\ref{stakeholder_identification}. Equally, based on the
uncertainty that we observed during the workshop sessions when
discussing priorities (see section~\ref{priorities}), participants may
have requested this guidance because they wanted their organizations
to relieve them of the responsibility of making prioritization
decisions themselves.

Participants acknowledged that organization-wide guidance might be
difficult to provide (despite the resources available to larger
organizations) given the diversity of AI systems, use cases, and
deployment contexts. That is, \textit{``[fairness] is hard to provide
  company-wide guidance on because it does vary on a case by case
  basis of what is the task and scenario and, like, how is the UI
  structured''} (P13, T2). One approach suggested by participants was
for organizations to create guidance (such as a list
of demographic groups that might be most at risk of experiencing poor
performance) that could then be adapted, supplemented, or prioritized
by teams based on their specific circumstances:
\begin{quote}
    \textit{I think that list should be global across all products, but then, like, morphological differences or a difference in skin color may not directly affect the way that you interact with the [product] UI but it could change your lived experience and you could still be served differently. So, like, some of them will be more relevant to certain services than others and that's where the prioritization should happen on the team's parts, but we need to be told, like, `Here's a list. And if you don't have coverage or you haven't at least considered and prioritized these then you're missing a requirement.'} (P13, T2)
\end{quote}

This desire to be given organization-wide guidance (especially
guidance relating to direct stakeholders and demographic groups) that
could be tailored to teams' specific circumstances was echoed by other
participants. Another participant on a team developing several related
AI systems said:\looseness=-1
\begin{quote}
    \textit{It makes sense to think about these factors and groups
      globally, and then to sort of double check for each one of the
      model types [...]  `Okay, well, how would that actually get
      reflected in all these model types? And would it be equally so
      in all of the model types?} (P7, T4)
\end{quote}

Participants also expressed desires for organization-wide strategies
for collecting datasets with which to conduct disaggregated
evaluations. Several participants pointed out that discussions about
data collection are \textit{``going to be something that comes up
  consistently organizationally for [our company]. What is that data
  and how do we balance that data collection [with privacy]?''} (P9,
T1). Participants felt that their organizations should help teams
understand how to collect demographic data, addressing challenges at
an organizational level, rather than at a team level. One participant
explicitly suggested centralizing data collection efforts, saying that
they wanted to:
\begin{quote}
    \textit{Give sort of a vision pitch of there being a group within
      [the company] who is responsible for a data warehouse or data
      clearinghouse for datasets that other teams can use to assess
      the fairness and bias in their algorithms} (P9, T1).
\end{quote}

However, much like organization-wide guidance, organization-wide
strategies for collecting datasets (including centralized data collection
efforts) might be difficult to establish given the diversity of AI
systems, use cases, and deployment contexts. Moreover, the economies
of scale that motivate the deployment of AI systems to new
geographic contexts based on strategic market tiers (as described in
section~\ref{priorities_business}) may lead to homogenized
understandings of demographic groups that may not be reflective of all
geographic contexts.


\subsubsection{Resources}
\label{resource_advocacy}

In addition to wanting their organizations to provide guidance about
designing and conducting disaggregated evaluations, participants
described needing to advocate for resources (e.g., money, time,
personnel) for designing and conducting disaggregated evaluations. The
need for resources to enable teams to prioritize fairness work
has been identified in prior research
\cite[e.g.,][]{madaio2020co,rakova2020responsible,metcalf2019owning}. Here,
we offer additional evidence and identify tensions in organizational
processes for advocating for resources.
Participants told us how,
\textit{``the main thing just comes down to funding [...] [VPs] should
  fund it and then it becomes much easier''} (P17, T8). One
participant highlighted the need for resources to
\textit{``institutionaliz[e] the practices that we espouse''} (P7,
T4). Participants on other teams agreed, saying \textit{``we're trying
  to figure out where the bottlenecks are. What kind of resourcing you
  would need to achieve an outcome like that, but those outcomes
  ideally should not be bottom up at [our team] level. They should be
  top down, from [our organization]''} (P3, T2).

Many participants discussed their teams' processes for identifying
such resource bottlenecks and how they used the resulting
conversations to advocate for resources to support their fairness
work. Without those resources, \textit{``we can only do so much with
  what we're given''} (P6, T1) and, as another team told us,
\textit{``if anybody wants us to do additional testing, which requires
  additional data gathering or labeling of existing data, right now we
  don't have any budget set aside for that, so we need to proactively
  plan for that''} (P7, T4). In other words, despite organizations'
stated fairness principles and practitioners' best intentions when designing
disaggregated evaluations, the reality of budgets for collecting
datasets, as well as budgets for other activities (such as engaging
with direct stakeholders and domain experts) constrain what teams are
able to achieve.

Over and over, participants told us that their organizations' business
imperatives dictated the resources available for their fairness work,
and that resources were made available only when business
imperatives aligned with the need for disaggregated evaluations. As
one participant told us:\looseness=-1
\begin{quote}
    \textit{If we bring these concerns up with management which is
      focused on business goals, they are not very interested in
      having this conversation. I guess it's a trickle down
      effect. When we see that the business stakeholders at the top
      have really adopted this as important, then it trickles down and
      that is something which becomes important to our immediate
      managers as well. These are all very consumer- and
      business-focused companies. So obviously, if it hurts their
      interests, we see what happens to AI researchers, ethics teams
      and the like... but if there's executive buy-in, then this could
      trickle down to all levels, for all sizes of organizations,
      because these are primarily capitalistic, economic ventures.}
    (P30, T6)
\end{quote}

The tension between business imperatives and fairness work (including
disaggregated evaluations) arose throughout the workshop sessions, but
was explicitly identified by one participant (P30) in reference to the
firing of Google's AI ethics research team
co-leads.\footnote{https://www.fastcompany.com/90608471/timnit-gebru-google-ai-ethics-equitable-tech-movement}
Along these lines, another participant that participated in a phase one
interview, but whose team declined to participate in the subsequent
workshop sessions, said:
\begin{quote}
    \textit{For someone who's looking at like `OK, I want to increase the market share production by X.' You're not moving the needle there, so then like how would you prioritize [disaggregated evaluations]? [...]
    Unless you have a clear buy-in, clear funding, it's very hard to get these things prioritized right?} (P17, T8)
\end{quote}

Participants told us how they would advocate for resources for
fairness work with their leadership, but explained that they would
need evidence that their AI systems caused fairness-related harms in
order to convince their leadership to actually provide
resources. Several teams shared frustrations about this vicious cycle
of needing evidence of performance disparities to secure resources to
design and conduct disaggregated evaluations---the same evaluations
that would provide the requested evidence of performance disparities.
At the end of the second workshop session with one team, the PM
mentioned that ``\textit{we should have a meeting where we could
  advocate for some of these things with [our VP] and make it really
  clear that he might have to shake some trees and change some
  minds}'' (P9, T1). This participant described how they wanted to
``\textit{make evidence a criterion for supporting our assessment
  prioritization}'' (P9, T1), but for them, that was more of an
aspirational goal than a reality:\looseness=-1
\begin{quote}
    \textit{I do think that I like to go into discussions like that
      with the carrot and stick approach, saying `Here are the things
      that can go wrong if we don't do this or that we should be
      worried about.' I feel like we've got an okay handle on
      that. But not for, like, `Here's some examples where it's really
      gone right where, like, look, we've quantified some differences
      between groups. Here's a case study where it has worked. This is
      how they just collected a little more data and adjusted the
      training process to account for that'} (P9, T1).
\end{quote}

That is, for this participant, and for others that were discussing how
to advocate for resources for designing and conducting disaggregated evaluations, the
\textit{``dream state''} (P9, T1) was to have case studies or examples
of AI systems where disaggregated evaluations worked---that is, where
teams had found quantified evidence of performance disparities,
enabling them to then mitigate those disparities. However, although
this participant's team was aware of potential fairness-related harms caused by their AI system, they were concerned that without additional
evidence, this would not be sufficient for them to secure the
resources they needed to design and conduct disaggregated evaluations.\looseness=-1

\section{Discussion}

Prior research has proposed disaggregated evaluations of AI systems as
a way to uncover performance disparities between demographic groups
\cite[e.g.,][]{barocas2021designing}. However, technology work
practices are shaped by the organizational contexts within which practitioners are embedded
\cite[e.g.,][]{passi2019problem,passi2020making,wolf2018participating},
and fairness work is no exception in this regard
\cite[e.g.,][]{madaio2020co,rakova2020responsible,metcalf2019owning,miceli2020between}. Indeed,
some research has suggested that first-party assessments of the
impacts of AI systems may be particularly susceptible to
organizational factors \cite{moss2021assembling}. We therefore used a
process for designing disaggregated evaluations that we adapted from
Barocas et al.~\cite{barocas2021designing} to explore practitioners'
existing processes and challenges when designing disaggregated
evaluations of their AI systems (RQ1), their needs for organizational
support (RQ2), and how their processes, challenges, and needs for
support are impacted by their organizational contexts (RQ3). We find
that practitioners face challenges when choosing performance metrics,
identifying the most relevant direct stakeholders and demographic
groups on which to focus (due to a lack of engagement with
direct stakeholders or domain experts), and collecting datasets with
which to conduct disaggregated evaluations. We discuss how the
heuristics that teams use to determine priorities for assessing the
fairness of AI systems may compound existing inequities. We find that
practitioners want their organizations to provide guidance on
identifying direct stakeholders and demographic groups and strategies
for collecting datasets, and we identify tensions in organizational
processes for advocating for resources for designing and conducting
disaggregated evaluations.\looseness=-1

In the rest of this paper, we discuss some implications of these findings.
We
extend prior research on the ways in which organizational factors,
including organizational cultures and incentives, impact fairness work
by discussing the implications of the ways in which practitioners' decisions
when designing disaggregated evaluations are influenced by business
imperatives such as tier-based deployment approaches and a tendency to
prioritize customers over marginalized groups. We also discuss how the
drive to deploy AI systems at scale impacts disaggregated
evaluations. A lack of processes for understanding what
marginalization means in different
geographic contexts causes practitioners to draw on the personal
experiences and identities represented on their teams or to use data
from their teams or organizations. These approaches may compound
existing inequities given the homogeneous demographics of many AI
teams \cite{west2019discriminating}.

\subsection{Implications of business imperatives that shape disaggregated evaluations}

Traditions of user-centered design from the HCI community have long
grappled with the political implications of the question, ``For whom
do computational systems (fail to) work?'' Now, as the HCI and AI
communities develop tools and practices to support practitioners in
identifying, assessing, and mitigating fairness-related harms caused
by AI systems, we must grapple with the political implications of who
is involved in fairness work. Prior research has pointed out that
incentives to ship AI products and services quickly may be at odds with
the slow and careful nature of fairness work
\cite[e.g.,][]{madaio2020co,rakova2020responsible}. In this paper, we
find that business imperatives shape decisions made by practitioners'
when designing disaggregated evaluations, including decisions about
performance metrics, direct stakeholders and demographic groups, and
datasets.

As Suchman pointed out, the term user ``opens out, on closer
inspection, onto an extended field of alliances and contests''
\cite{suchman2002located}. Popular approaches to identifying
stakeholders from the HCI and design communities have expanded and
problematized who is involved in technology development from users to
stakeholders more generally, offering methods
\cite[e.g.,][]{yoo2010understanding,yoo2013value,forlizzi2018moving},
theories \cite[e.g.,][]{friedman2017survey,friedman2006watcher}, and
frameworks \cite[e.g.,][]{suresh2021beyond} for identifying
stakeholders and understanding their values. In practice, the
approaches that our participants took\footnote{It is important to note
  that what we observed during the workshop sessions could not capture
  the entirety of teams' efforts around engaging with direct
  stakeholders. Whenever possible, we therefore asked participants to
  describe their typical processes for engaging with direct
  stakeholders. Our findings reflect the engagements that they
  described.} to identifying direct stakeholders and demographic
groups appear to have more in common with approaches for identifying
stakeholders found in business operations and management research
\cite[e.g.,][]{mitchell1997toward,de2017reviewing} than with
approaches found in the HCI or design communities. In business
operations (see \citet{de2017reviewing} for a review), stakeholders
are identified and involved based on either an instrumental rationale
(i.e., the belief that involving stakeholders will improve
organizations' performance or profitability) or a moral rationale
(i.e., the belief that involving stakeholders is the right thing to
do). For many of our participants, the business imperatives that they
described shaping their decisions appear to fit within instrumental
approaches to identifying stakeholders, where high-value customers (or
other stakeholders with the potential to improve organizations'
performance or profitability) and stakeholders in higher-tier markets are prioritized
over other stakeholders. Indeed, such priorities should make us
skeptical that organization-wide guidance on identifying direct
stakeholders and demographic groups or organization-wide strategies
for collecting datasets will actually reflect the needs of
marginalized groups.\looseness=-1

More generally, fairness work that ignores the role played by business
imperatives may inadvertently compound existing inequities. For
example, strategic market tiers that inform deployment schedules may
lead to disaggregated evaluations that follow similar approaches, in
turn resulting in the deprioritization of direct stakeholders and
demographic groups in lower-tier markets. Without processes for
engaging with direct stakeholders or domain experts, practitioners draw on their
personal experiences and identities, their perceptions about
fairness-related harms (including the perceived severities of those harms), and
even their own data---all workarounds that may perpetuate majoritarian
structures of marginalization.\looseness=-1

Within HCI, recent work on disability justice has critiqued the
rhetoric around designing with empathy for marginalized groups, rather
than involving them in the development lifecycle
\cite{bennett2019promise, costanza2020design}. By relying on the
personal experiences and identities represented on AI teams,
practitioners may overlook fairness-related harms, especially given
the homogeneous demographics of many such teams
\cite{west2019discriminating}. One way to mitigate this is to recruit
practitioners from a wider range of backgrounds, including from
demographic groups that are currently under-represented in
AI. Although having more diverse AI teams is critical, the ``politics
of inclusion'' \cite{young2011justice, dixon2020data} of relying on
marginalized practitioners may not be sufficient to effect systemic
change given the dominant structural forces
\cite[e.g.,][]{ray2019theory} that might lead to those practitioners
being ignored, tokenized, or fired \cite{dixon2020data,
  young2011justice,hoffmann2020terms}. Recent calls to foster greater
engagement with affected communities when developing sociotechnical
systems may serve as a counter to business imperatives
\cite[e.g.,][]{costanza2020design,tran2019gets}, although it is
important to be wary about extractive, tokenistic approaches
\cite[e.g.,][]{sloane2020participation, asad2019academic,
  arnstein1969ladder}, which can unfairly burden members of
marginalized groups \cite{pierre2021getting}.\looseness=-1

\subsection{Implications of deploying AI systems at scale}

Many of our participants reported challenges relating to the scale at
which AI systems are deployed.\footnote{We note that our focus on
  geographic scale is complementary to recent research grappling with
  other notions of scale (such as the size of training datasets
  \cite{bender2021dangers} or the number of users
  \cite[cf.][]{gillespie2020content}) which are out of scope for this
  paper.} Participants on every team shared that they felt pressured
to expand deployment to new geographic contexts, and we saw the impact
of these pressures on nearly every decision made when designing
disaggregated evaluations.

Critical scholarship, including recent research in CSCW
\cite[e.g.,][]{tsing2012nonscalabilitythe,hanna2020against,lempert2016scale,rossitto2020reconsidering},
has explored how scale is not simply an objective property of
sociotechnical systems---for example, a quantified accounting of the
number of inputs or users that a system has, or a tally of the
different contexts in which a system is deployed. Indeed, as in our
findings, it is often a non-trivial matter to even identify what
context means for an AI system's deployment---does it refer to
countries? geographic regions? strategic market tiers? Instead, we
draw on the concept of scalar thinking from Tsing and others
\cite{tsing2012nonscalabilitythe,lempert2016scale} to refer to the
discourses of scale that motivate and enable AI systems to
``proliferate across contexts and over time''
\cite{rossitto2020reconsidering}. These discourses are part of larger
forces within technology development that valorize the scalability of
systems as a precursor to venture capital investment
\cite{hanna2020against}. \looseness=-1

The expansionist rhetoric around AI systems raises serious questions
for fairness work. To what extent can disaggregated evaluations
conducted by teams embedded within technology companies reveal
fairness-related harms caused by AI systems that are deployed to
multiple geographic contexts? This question has a long historical
resonance. Suchman argued that we must attend to the specificities of
place in technology development practices, including its micropolitics
and cultural imaginaries, to avoid the reproduction of ``neocolonial
geographies of center and periphery'' (and, more generally, to resist
such neat binaries of center and periphery, local and global)
\cite{suchman2011anthropological}. More recently, Sloane et al.
argued that the ever-increasing drive to deploy AI systems at scale
may be fundamentally at odds with calls to involve direct stakeholders
in the development of those systems
\cite{sloane2020participation}. Our findings reveal implications of
this tension, as many participants reported deploying AI systems in
geographic contexts for which they have no processes for engaging with
direct stakeholders or domain experts. As a result, they noted that they struggle to
understand the impacts of their systems in those contexts and,
therefore, how best to collect datasets with which to conduct
disaggregated evaluations. The consequences of this can be dire. As
leaked documents have
revealed,\footnote{https://www.theatlantic.com/ideas/archive/2021/10/facebook-failed-the-world/620479/}
Facebook's hate speech detection system disproportionately failed to
identify hate speech in languages other than English---a problem
exacerbated by a lack of resources to support this work.\footnote{See
  also Timnit Gebru's ICLR 2021 talk:
  \url{https://iclr.cc/virtual/2021/invited-talk/3718}}\looseness=-1

The scalability of sociotechnical systems is often thought of in terms
of the technical feasibility of deploying those systems to new
contexts, relying on standardized technical infrastructures (e.g.,
cloud computing) and technologies of standardization like containers
(in both the physical and digital senses of the term container)
\cite[cf.][]{jackson2015standards,timmermans2010world}. One
implication of our findings, however, is that human-centered
requirements (such as, but not limited to, engagement with direct
stakeholders) should be considered on an equal footing with technical
feasibility. Tsing articulated a theory of nonscalability, or how to
think about systems that change in response to local conditions as
they grow \cite{tsing2012nonscalabilitythe}. Therefore, if scalable
projects, broadly construed, are those that are able to grow without
changing, then nonscalability suggests that (in this case) development
practices for AI systems should change when expanding into new
contexts, so as to respond to local conditions. These changes may
involve data collection, model retraining, and, crucially,
involving direct stakeholders in the development lifecycle (including
in discussions about whether proposed AI systems should exist at all
\cite[cf.][]{baumer2011implication,barocas2020not}).\looseness=-1

Future work should consider what it means for AI systems to be
nonscalable
\cite[cf.][]{gillespie2020content,tsing2012nonscalabilitythe}, so as
to resist the ``portability trap'' of believing that one can simply
port AI systems developed in and for one context to others
\cite{selbst2019fairness}. This may involve developing AI systems (and
disaggregated evaluations) in ways that are responsive to local values
and norms, as in recent research to re-imagine the fairness of AI
systems in India \cite{sambasivan2021re}, research that
has highlighted the risks of algorithmic colonization
\cite{birhane2020algorithmic} and efforts toward decolonial AI
\cite{mohamed2020decolonial}, or, more generally, approaches for
``designing for the pluriverse'' \cite{escobar2018designs}. However,
approaches to the development of AI systems that are responsive to local
conditions will likely require slow and careful work that is
fundamentally at odds with the \textit{``pedal [...] to the metal''} (P7, T4) approach
incentivized by business imperatives.\looseness=-1

\subsection{Limitations}
\label{limitations}

Thirty-three practitioners took part in our study, from ten teams
responsible for developing AI products or services at three technology
companies. Ideally, we would have been able to provide more
information about these teams and companies, but participants agreed
to participate on the condition that details about their companies and
their products and services would be abstracted to preserve
anonymity. Our recruitment strategy (i.e., direct emails and posts on
message boards related to the fairness of AI systems) and our
positionality as researchers living and working in the
U.S., primarily working in industry, with backgrounds in AI and HCI may have limited the range of
practitioners that agreed to participate, suggesting that future
work on this topic should take a broader recruitment strategy so
as to recruit participants from a wider range of teams and
companies. Our positionality
may also have shaped how we approached our research questions, data
collection, and data analysis. Future work conducted by people with
other backgrounds and from other contexts, including people outside
of academia or industry (e.g., community groups, government agencies,
civil society organizations), is therefore crucial.\looseness=-1

In addition, most participants in the workshop sessions
were members of teams developing language technologies (six of the
seven teams), although we we did conduct interviews with PMs from
three additional teams working on other AI products and services that
declined to participate in the subsequent workshop sessions. Although
our findings suggest implications for teams developing AI systems
other than language technologies and for companies beyond the three
that participated, they may not be generalizable to all teams and
companies. Future work should therefore involve larger-scale studies
to validate or refute our findings, and, especially, studies involving
practitioners whose organizations require them to design and conduct
disaggregated evaluations rather than doing so voluntarily. Finally,
our findings are based on the interviews and workshop sessions, which
were conducted over a limited period of time. Future work should
include longitudinal studies such as observational studies of teams
designing and conducting disaggregated evaluations as part of their
work practices.\looseness=-1

\subsection{Conclusion}

As researchers and practitioners develop new tools and practices for
identifying, assessing, and mitigating fairness-related
harms caused by AI systems, it is critical to understand how these
tools and practices are actually used. In this paper, we focus on one
such practice: disaggregated evaluations of AI systems, intended to
uncover performance disparities between demographic groups. Via
semi-structured interviews and structured workshops with AI
practitioners at multiple companies, we identify impacts on fairness
work stemming from a lack of engagement with direct stakeholders or domain experts,
business imperatives that prioritize customers over marginalized
groups, and the drive to deploy AI systems at scale. Specifically, we
find that practitioners face challenges when choosing performance
metrics, identifying the most relevant direct stakeholders and
demographic groups on which to focus, and collecting datasets with
which to conduct disaggregated evaluations. These findings suggest the
need for processes for engaging with direct stakeholders and domain experts prior to
deployment to new geographic contexts, as well as counterbalances to
business imperatives that can lead to pressures to deploy AI systems
before assessing their fairness in contextually appropriate ways.\looseness=-1

\bibliographystyle{ACM-Reference-Format}
\bibliography{cscw_fairness}

\received{July 2021}
\received[revised]{November 2021}
\received[accepted]{November 2021}
\end{document}